\documentclass[10pt,twocolumn,letterpaper]{article}

\usepackage{wacv}              

\usepackage{verbatim}
\definecolor{wacvblue}{rgb}{0.21,0.49,0.74}
\usepackage[pagebackref,breaklinks,colorlinks,allcolors=wacvblue]{hyperref}

\def\wacvPaperID{389} 
\def\confName{WACV}
\def\confYear{2026}

\title{EntropyGS: An Efficient Entropy Coding on 3D Gaussian Splatting}

\author{Yuning Huang, Jiahao Pang,  Fengqing Zhu, Dong Tian}
\begin{document}
\maketitle
\begin{abstract}
As an emerging novel view synthesis approach, 3D Gaussian Splatting (3DGS) demonstrates fast training/rendering with superior visual quality. The two tasks of 3DGS, Gaussian creation and view rendering, are typically separated over time or devices, and thus storage/transmission and finally compression of 3DGS Gaussians become necessary.
We begin with a correlation and statistical analysis of 3DGS Gaussian attributes.
An inspiring finding in this work reveals that spherical harmonic AC attributes precisely follow Laplace distributions, while mixtures of Gaussian distributions can approximate rotation, scaling and opacity. 
Additionally, harmonic AC attributes manifest weak correlations with other attributes except for inherited correlation from a color space.
A factorized and parameterized entropy coding method, EntropyGS, is hereinafter proposed.
During encoding, distribution parameters of each Gaussian attribute are estimated to assist their entropy coding. 
The quantization for entropy coding is adaptively performed according to Gaussian attribute types.
EntropyGS demonstrates about $30\times$ rate reduction on benchmark datasets while maintaining similar rendering quality compared to input 3DGS data, with a fast encoding and decoding time. 
\end{abstract}

\section{Introduction}

Novel View Synthesis (NVS) is a computer graphics task aimed at generating images from novel viewpoints in applications like virtual reality, 3D reconstruction, and gaming.
By introducing Gaussians as primitives, 3D Gaussian Splatting (3DGS)~\cite{3DGS} effectively represents both the geometry and texture information of a 3D scene. 
A Gaussian primitive is composed of position, rotation, opacity, scaling, and spherical harmonics to capture view-dependent color.
With its highly parallelized design and advancements in GPU technology, 3DGS achieves significantly faster training and rendering times than the Neural Radiance Field (NeRF)~\cite{NeRF} framework, while delivering competitive visual quality.
However, the high data rate of Gaussian primitives poses a significant challenge to the transmission/storage of 3DGS model, when its training and rendering are separated over time or devices.

\label{sec:EntropyGS_intro}
\begin{figure}[t]
\centering
\includegraphics[width=0.47\textwidth]{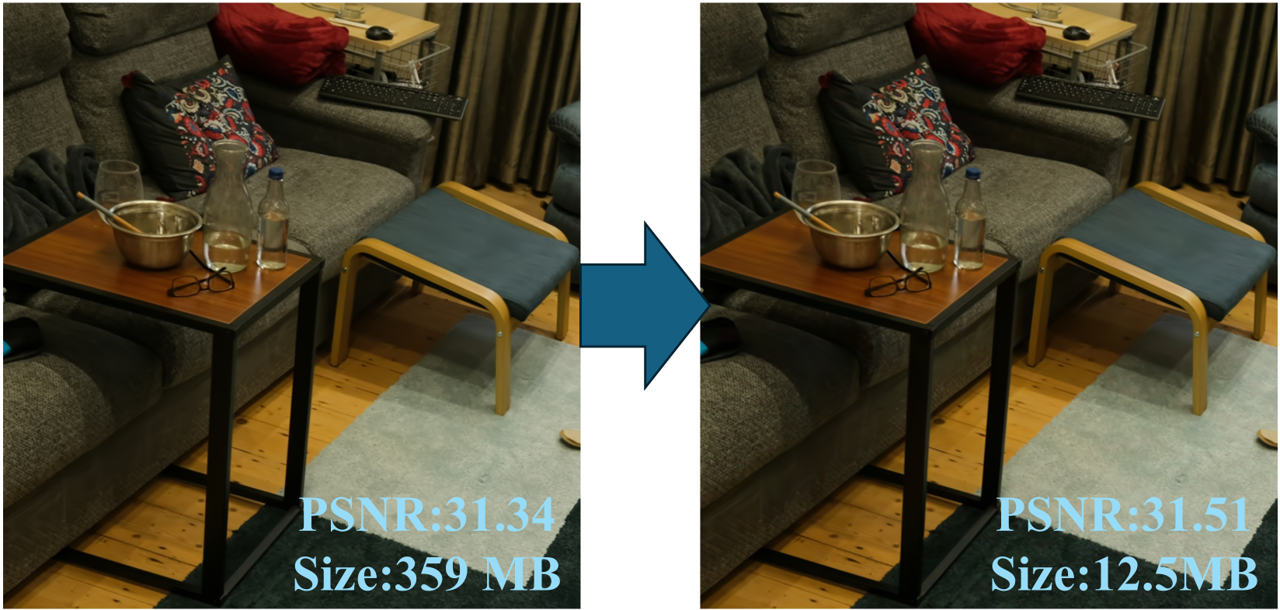}
\caption{Our proposed EntropyGS offers a noticeable memory reduction of approximately $30\times$ compared to 3DGS~\cite{3DGS}, while maintaining high rendering quality. Additionally, the encoding and decoding times are kept under 5 seconds for scenes such as ``Room'' from the Mip-NeRF 360 dataset~\cite{mipnerf}.} 
\label{fig: EntropyGS-head}
\end{figure}

The attempt toward efficient encoding of 3DGS into a bitstream could be differentiated depending on whether offline updating of 3DGS Gaussians is allowed. For applications where offline updating of Gaussians is feasible, a global optimization across Gaussian creation/updating and compression is applied~\cite{scaffoldgs,HAC,SOG1}. It provides more potential for compression while requiring additional offline processing to update Gaussians.
For applications where low latency is a priority or offline processing is not affordable, the Gaussian primitives, as a pre-generated 3DGS model, are encoded into a bitstream as they are. Hence, compression-specific studies with a limited scope have their own merit~\cite{LightGaussian}.
This is how our work is positioned.

Many neural network (NN) model compression techniques, e.g., model parameter quantization, pruning~\cite{structurepruning,softpruning} and knowledge distillation~\cite{kd1,kd2}, are leveraged for 3DGS compression.
However, unlike our proposal, in those works, neither the physical significance, nor the distinct statistical characteristics, of Gaussian attributes (position, rotation, .., spherical harmonics) is well counted.
Those 3DGS compression works are aligned with general NN model compression methodologies.

We start with the analysis of the statistical properties of 3DGS attributes.
Based on the theoretical induction and empirical verification, we reveal the weak correlation between spherical harmonic AC attributes, except for the inherited correlation in the color space.
The discovery provides a solid motivation for our proposed \emph{factorized} entropy coding design.

Additionally, for each attribute, we reveal that their statistics have
a strong coherence with a common distribution type, e.g., Gaussian or Laplace. 
It leads to the proposed entropy coding depending on \emph{statistical estimation} without counting on learning-based methods.
Consequently, we propose an entropy coding method for 3DGS, named as EntropyGS,
a \emph{factorized} and \emph{parameterized} design. It is novel and efficient with statistical coherence being fully explored for the first time in literature.

Main contribution of our work is summarized as follows:
\begin{enumerate}
\item {A factorized design is proposed to entropy encode 3DGS based on a correlation analysis of Gaussian attributes that indicates a weak correlation in particular between spherical harmonic AC components.}
\item {A parameterized design is further proposed using common distributions to govern Gaussian attributes, including rotation, scaling, opacity, spherical harmonic AC.}
\item{An adaptive quantization finally completes the design of entropy encoder EntropyGS for pre-generated 3DGS models with competitive rate-distortion performance.}
\end{enumerate}

\section{Related Works}
\label{sec:EntropyGS_relatedworks}
For different applications depending on whether offline updating Gaussian primitives is permitted, existing 3DGS encoding can be categorized into two classes: \emph{joint optimization and compression}, and \emph{compression-specific} methods.

Methods in the former class demonstrate more compression capability as they allow further end-to-end optimization. Methods in the latter class, however, implement low latency and no alteration of 3DGS representations.

\begin{figure*}[h]
    \centering
    \hfill
    \begin{subfigure}[b]{0.3\textwidth}
        \includegraphics[width=\textwidth]{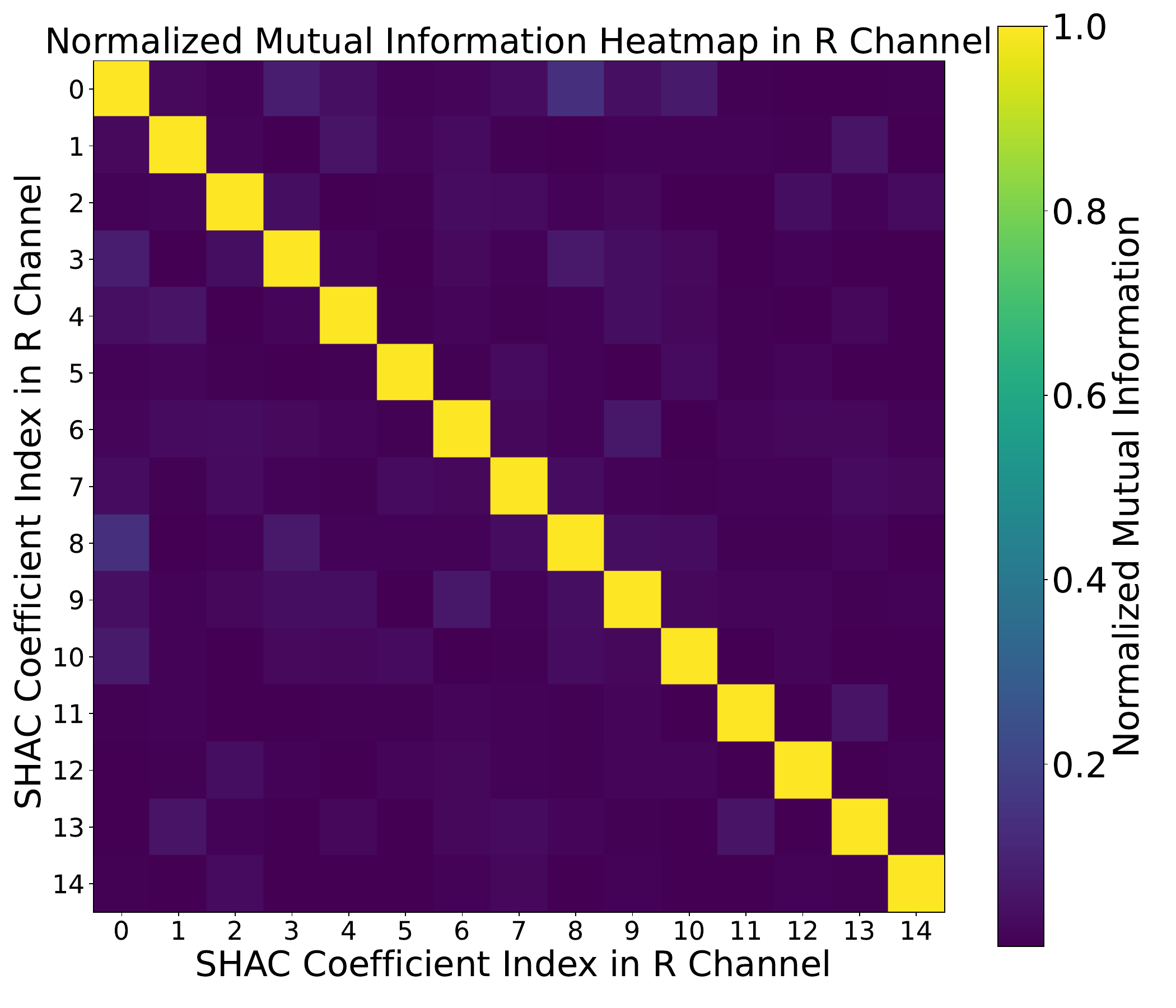}
        \caption{SHAC-SHAC, R}
    \end{subfigure}
    \hfill
    \begin{subfigure}[b]{0.3\textwidth}
        \includegraphics[width=\textwidth]{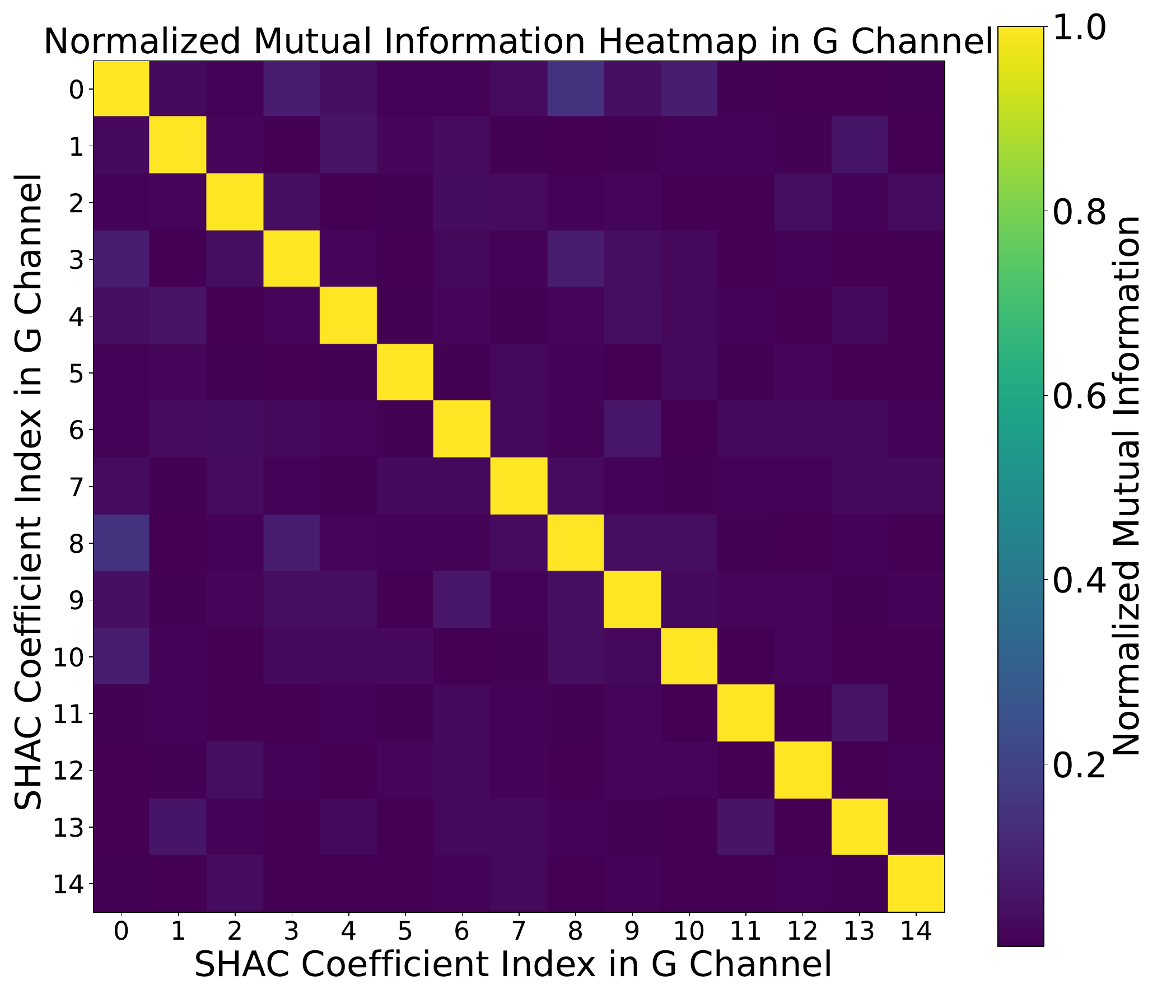}
        \caption{SHAC-SHAC, G}
    \end{subfigure}
    \hfill
    \begin{subfigure}[b]{0.3\textwidth}
        \includegraphics[width=\textwidth]{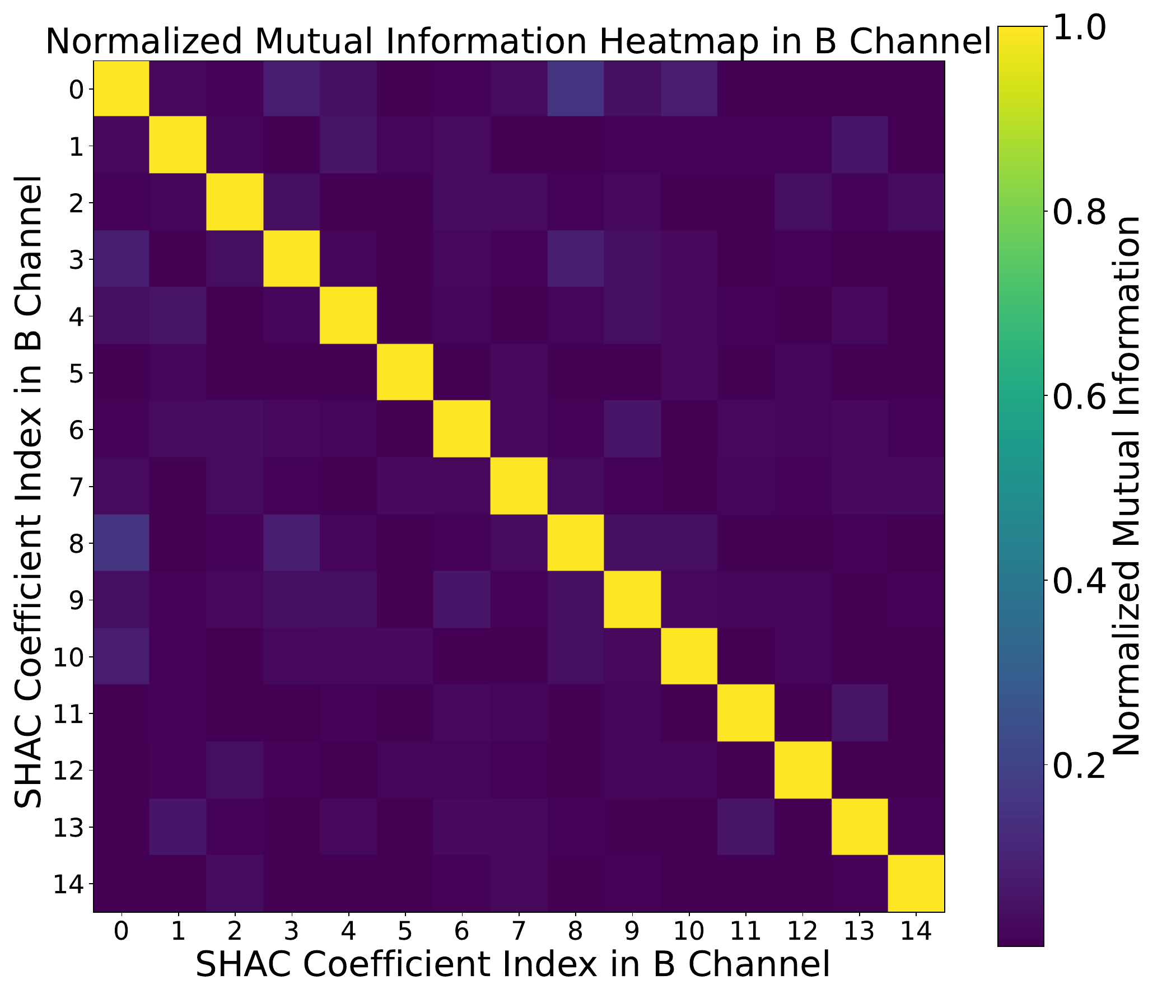}
        \caption{SHAC-SHAC, B}
    \end{subfigure}
    
    \vspace{0.1cm}
    
    \begin{subfigure}[b]{0.5\textwidth}
        \includegraphics[width=\textwidth]{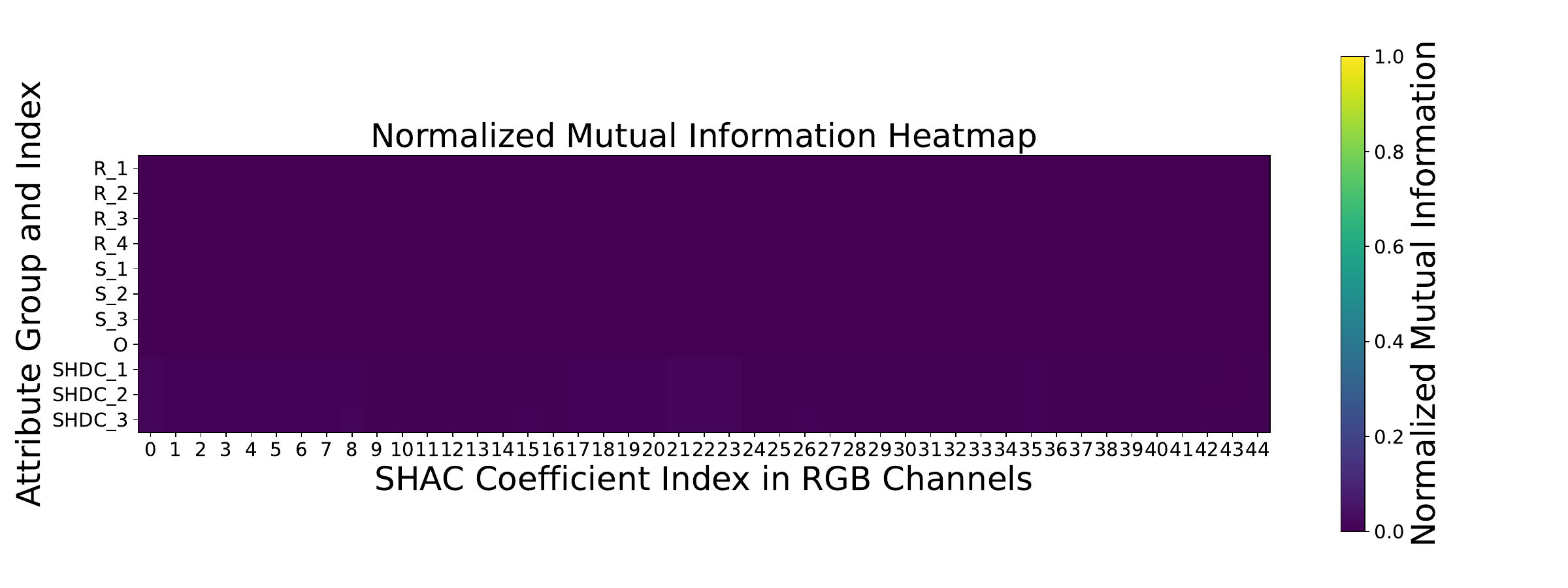}
        \caption{SHAC-(Rotation, Scaling, Opacity, SHDC)}
    \end{subfigure}

    \caption{Mutual Information heatmap of ``bicycle'' from Mip-NeRF 360 dataset. The first row shows the intra-correlation within SHAC, second row shows the inter-correlation between SHAC and the other Gaussian attributes}
    \label{fig:images1}
\end{figure*}
\subsection{Joint Optimization and Compression}

For methods in this category, the 3DGS Gaussian generation process is typically updated by counting the data rate required for 3DGS compression as an additional loss term.
In addition, they may alter the original Gaussian representation by augmenting it with other formats, for example, 3D mesh~\cite{meshGS} and even interpolators based on MLP~\cite{scaffoldgs}. Such modifications should be careful for applications such as scene editing and interactive rendering.

On top of ScaffoldGS~\cite{scaffoldgs}, an anchor-based representation instead of the original 3DGS representation is adopted in HAC~\cite{HAC} for a joint optimization and compression.
It introduces a compression framework designed to reduce the memory footprint of Scaffold-GS by introducing a binary hash grid and associated context-based modeling.

SOG~\cite{SOG1} proposes another modified Gaussian representation. The parameters of 3D Gaussian Splatting (3DGS) are arranged onto a 2D grid, then a joint optimization and compression is conceived that promotes local homogeneity. However, the sorting required for a 3D-to-2D mapping results in a high encoding time.

Pruning and subsequent optimization~\cite{trim,LightGaussian,mesonGS} are also commonly used techniques in existing works for reducing the rate of 3DGS. Various pruning strategies may be employed to decide which Gaussian primitives to be pruned, e.g., based on their impact on the rendering quality.
Such pruning is typically harmonized with 3DGS Gaussian generation, hence they fall in the study of joint optimization and compression.

\subsection{Compression-specific Study}
For scenarios where low latency is a priority, the generation of 3DGS model and its compression have to be disentangled. Compression-specific studies target to encode a pre-generated 3DGS model. The creation of 3DGS model is an upstream and offline task for compression-specific methods.

A representative work in LightGaussian~\cite{LightGaussian} introduces techniques used in neural network model compression, including vector quantization, knowledge distillation, to compress the original 3DGS model. C3DGS~\cite{compressedgs} proposes sensitivity-aware vector clustering with quantization-aware training. MesonGS~\cite{mesonGS} replaces rotation quaternion with Euler angles and applies region adaptive hierarchical transform (RAHT)~\cite{de2016compression} to encode key attributes.

Entropy coding is an essential step for data compression, as well as 3DGS compression.
HAC~\cite{HAC} estimates distribution parameters using an overfitted MLP to perform entropy coding for 3DGS model.

Our work falls into a study about efficient entropy coding. 
In contrast to the literature, our method employs a parameterized design based on the identified distribution models. 
The distinction of our method lies in the design based on statistical estimation that is much more lightweight than a learning-based method.


\section{Statistical Analysis of 3DGS Attributes}
This section provides an analysis of the 3DGS attributes, which inspires our proposal for an efficient 3DGS compression.
Gaussian attributes are partitioned into six distinct groups hereinafter: geometry, rotation, scaling, opacity, SHDC for first-order spherical harmonic coefficients, SHAC for higher-order harmonic coefficients.

\subsection{Correlation Analysis}
We characterize the correlation analysis into two categories: \emph{intra correlation} within SHAC coefficients and \emph{inter correlation} between SHAC and other attributes, including rotation, scaling, opacity, SHDC.

For intra correlation within the SHAC channels, the orthogonality of spherical harmonics ensures that each component (or coefficient) is independent by construction. For inter correlation, the relationship between SHAC and other attributes, no strong correlations are expected because of their distinct physical meaning.
In conclusion, mathematical independence theoretically results in zero intra correlation and conceptual independence leads to negligible inter-correlation.

Minor correlation may be introduced during the Gaussian creation process. However, both intra correlation and inter correlation are assumed to be minor in this work.

Note that correlation from color space will be carried over between corresponding spherical harmonic coefficients. It is subject to future work how to utilize them for entropy coding.

\textbf{Metric for measuring the correlation:}
We use \emph{Normalized Mutual Information (NMI)} as the metric to evaluate the correlation. NMI quantifies the amount of shared information between two variables, hence capturing the general correlation (both linear and non-linear). 

The Normalized Mutual Information (NMI) between two random variables \(X\) and \(Y\), given their pmf (histogram), can be computed using the following formula:
\[
\text{NMI}(X, Y) = \frac{2 \cdot I(X; Y)}{H(X) + H(Y)}
\]
\noindent where:
\begin{itemize}
\setlength{\itemindent}{1em}
    \item \(I(X; Y)\): The mutual information between \(X\) and \(Y\):
    \[
    I(X; Y) = \sum_{x \in X} \sum_{y \in Y} p(x, y) \log \frac{p(x, y)}{p(x)p(y)}.
    \]
    Here:
    \begin{itemize}
    \setlength{\itemindent}{1em}
        \item \(p(x, y)\): Joint pmf of \(X\) and \(Y\).
        \item \(p(x)\): Marginal pmf of \(X\), \(p(x) = \sum_{y} p(x, y)\).
        \item \(p(y)\): Marginal pmf of \(Y\), \(p(y) = \sum_{x} p(x, y)\).
    \end{itemize}

    \item \(H(X)\): The entropy of \(X\), defined as:
    \[
    H(X) = -\sum\nolimits_{x \in X} p(x) \log p(x)
    \]

    \item \(H(Y)\): The entropy of \(Y\), defined similarly:
    \[
    H(Y) = -\sum\nolimits_{y \in Y} p(y) \log p(y)
    \]
\end{itemize}
The result is a value between 0 and 1: \(\text{NMI}(X, Y) = 1\) indicates perfect correlation, \(\text{NMI}(X, Y) = 0\) indicates no mutual correlation.

\begin{figure*}[h]
    \centering
    \begin{subfigure}[t]{0.24\textwidth}
        \centering
        \includegraphics[width=\textwidth]{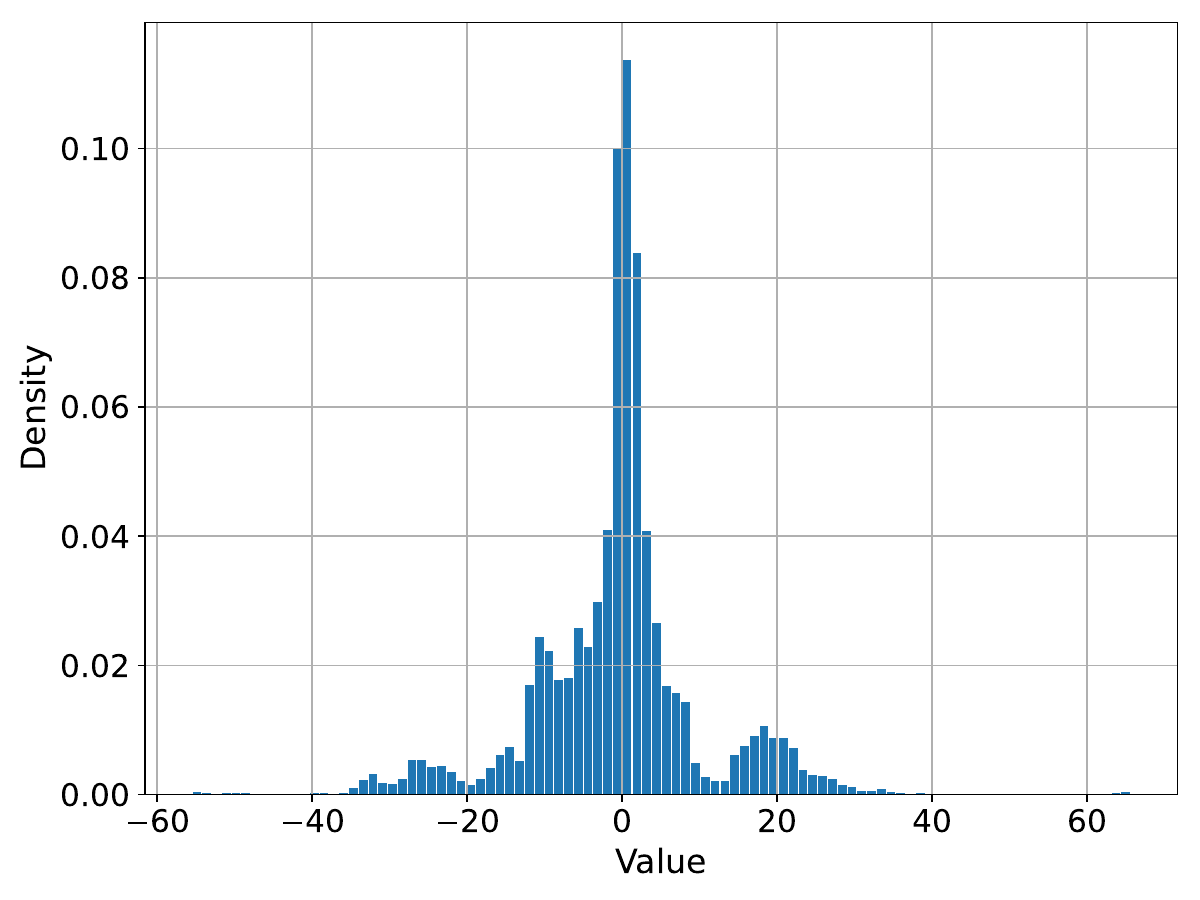}
        \caption{Geometry}

    \end{subfigure}
    \hspace{0.5cm}
    \begin{subfigure}[t]{0.24\textwidth}
        \centering
        \includegraphics[width=\textwidth]{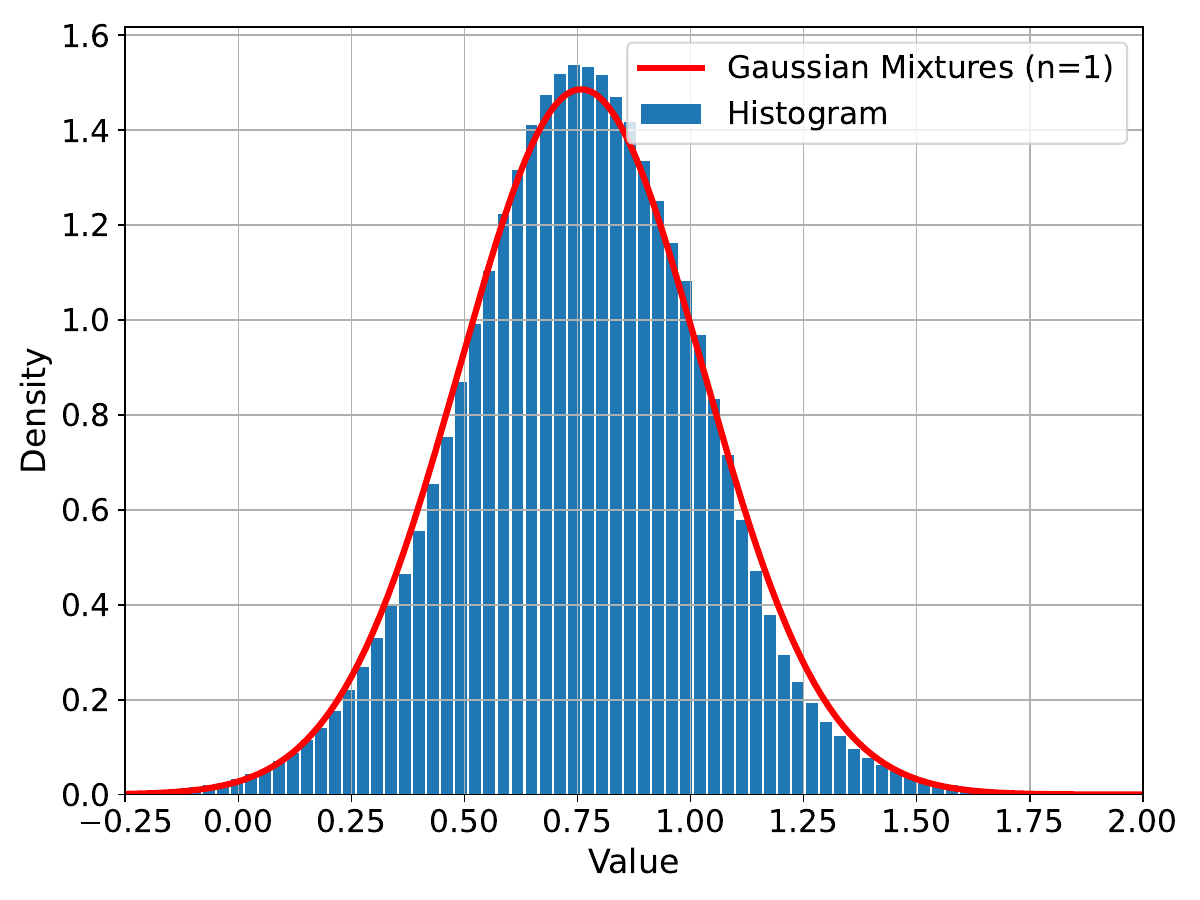}
        \caption{Rotation}
    \end{subfigure}
    \hspace{0.5cm}
    \begin{subfigure}[t]{0.24\textwidth}
        \centering
        \includegraphics[width=\textwidth]{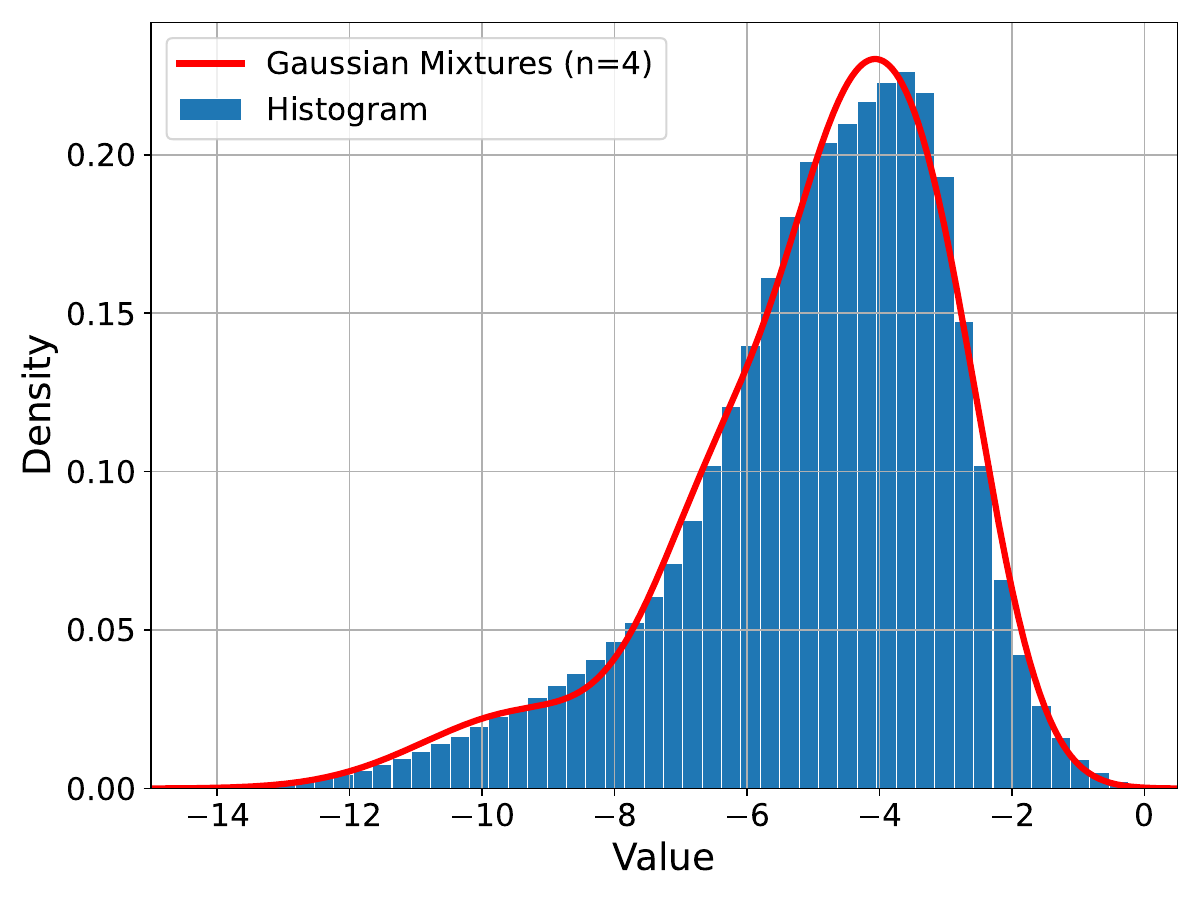}
        \caption{Scaling}
    \end{subfigure}


    \begin{subfigure}[t]{0.24\textwidth}
        \centering
        \includegraphics[width=\textwidth]{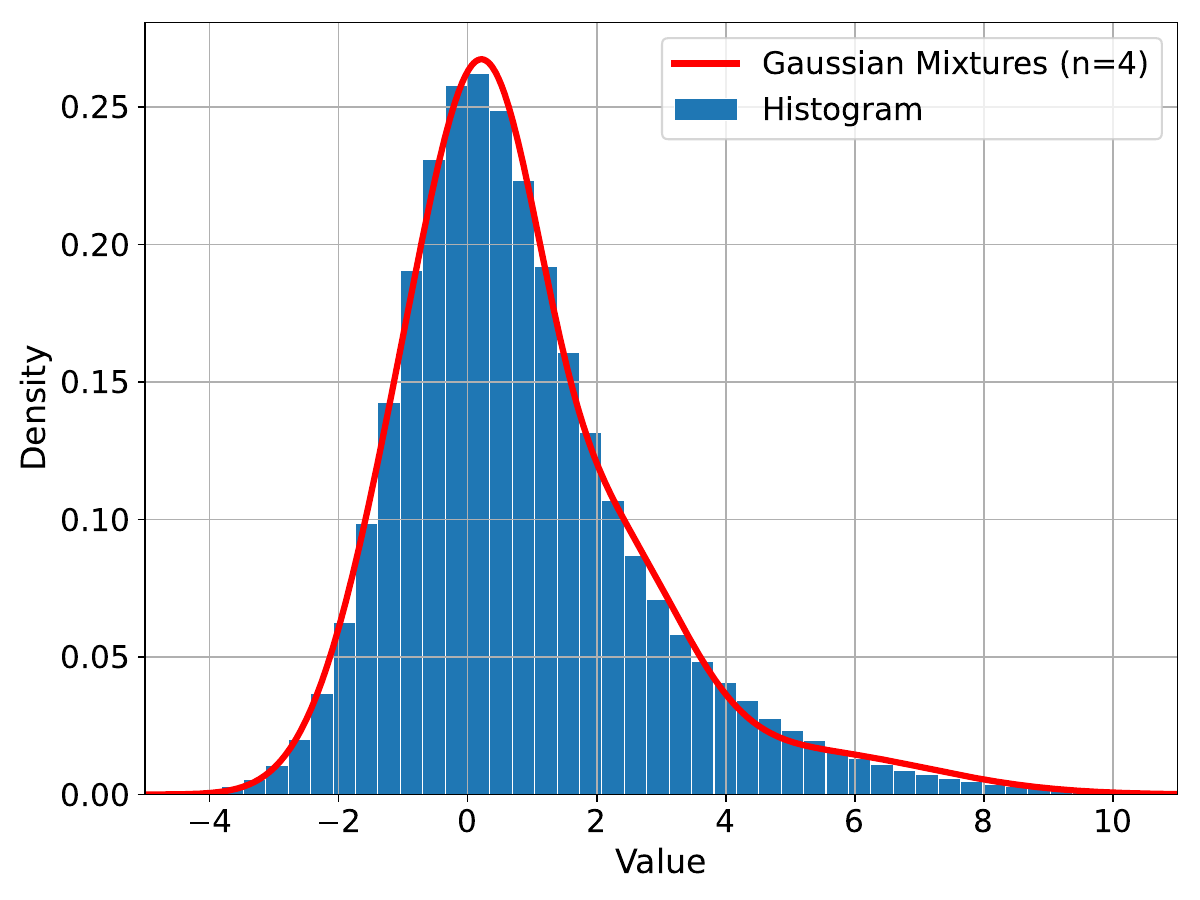}
        \caption{Opacity}
    \end{subfigure}
    \hspace{0.5cm}
    \begin{subfigure}[t]{0.24\textwidth}
        \centering
        \includegraphics[width=\textwidth]{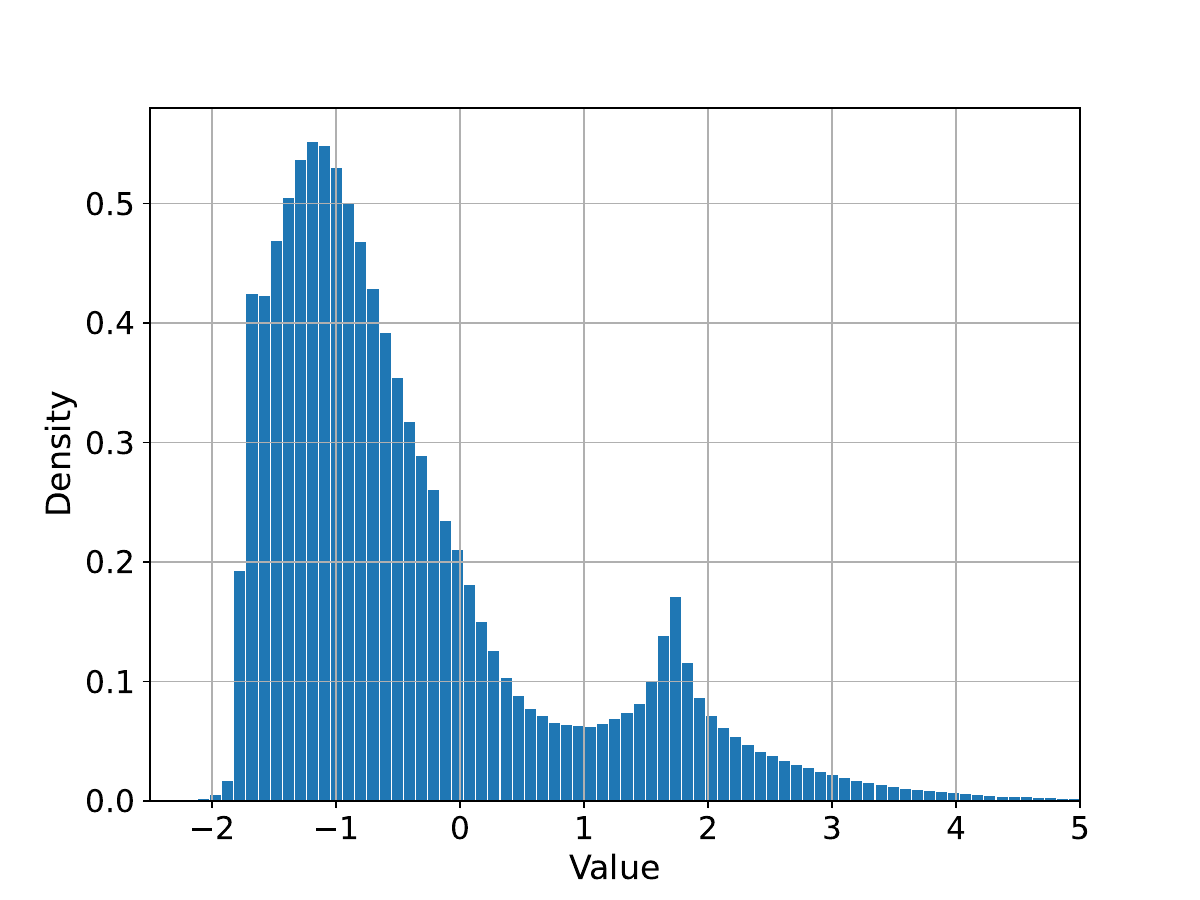}
        \caption{SHDC}
    \end{subfigure}
    \hspace{0.5cm}
    \begin{subfigure}[t]{0.24\textwidth}
        \centering
        \includegraphics[width=\textwidth]{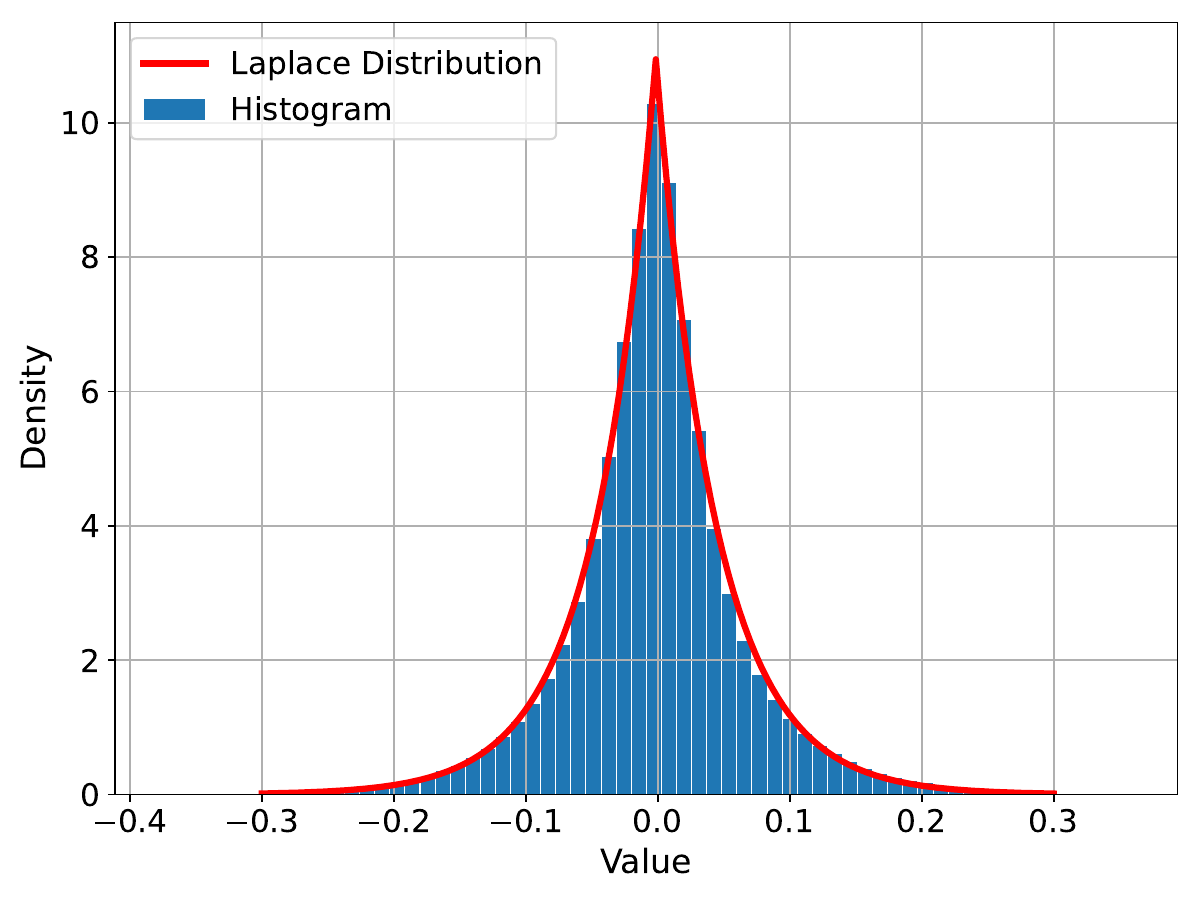}
        \caption{SHAC}
    \end{subfigure}
    \caption{Example histograms of Gaussian attributes and estimated distribution (red curve) from the ``Bicycle'' scene in the Mip-NeRF datasets. Each plot is related to one channel from the respective attribute group. We do not estimate the distribution for Geometry and SHDC. Note that similar statistical behavior is observed for other scenes across different datasets.}
    \label{fig: EntropyGS_comparison1}
\vspace{-0.5cm}
\end{figure*}

\begin{figure}[t]
\centering
\includegraphics[width=0.32\textwidth]{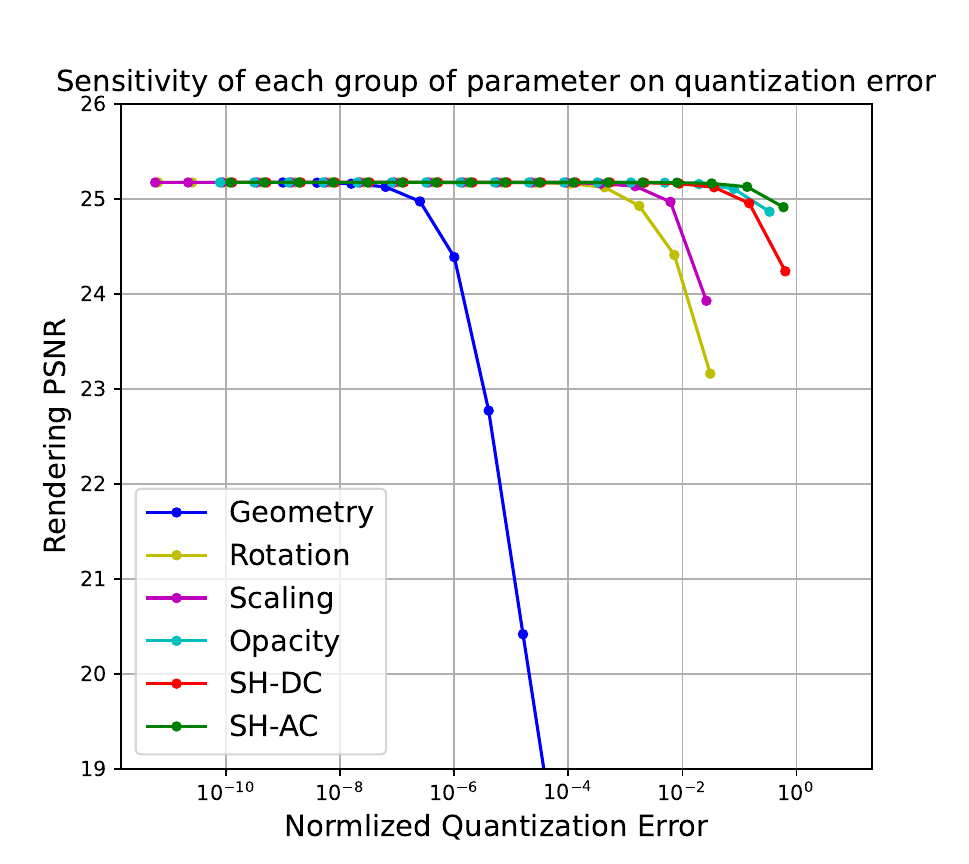}
\caption{Sensitivity comparison}
\label{fig: EntropyGS_sense}
\vspace{-0.5cm}
\end{figure}

The normalized mutual information heatmap for a representative scene is shown in Fig.~\ref{fig:images1}. From the first row of the figure, a weak intra-correlation is observed. The second row of the figure has verified that no inter-correlation exists between SHAC and other attributes. Please note that all tested scenes have shown the same trend, please see the full results for other scenes in the supplementary.

The observed weak correlation provides a solid support for a factorized entropy model design, where we can treat each channel of Gaussian attributes independently. Such design presents high efficiency, flexibility, and can be easily optimized for a parallel computation platform to achieve fast encoding and decoding time.

During analysis, we find the correlation between Gaussian attributes is stronger for close Gaussian primitives, this is reasonable as local similarity is expected. However, efficiently incorporating this correlation is challenging (KNN needed) and would break the factorized design of our entropy model; hence, we consider it out of scope.

\subsection{Governing Distributions for 3DGS attributes}
We hereby analyze the statistical behavior of 3DGS attributes, \textit{i.e.}, geometry, rotation, scaling, opacity, and both the DC and AC components of the spherical harmonics (SH). Sample histograms and estimated distributions are presented in Fig.~\ref{fig: EntropyGS_comparison1}. Beyond this specific example, we observe consistent statistical behavior across all scenes in the benchmarking datasets.

Empirical results across diverse datasets reveal distinct statistical characteristics for each attribute group:
\begin{itemize}
\setlength{\itemindent}{1em}
\item \textbf{Geometry}: Exhibits a highly dynamic range and sparsity, with no clear alignment to commonly-used distributions;
\item \textbf{Rotation/Scaling/Opacity}: Generally follows a simple Gaussian Mixture Model with fewer than four components, and in many cases can be approximated by a single Gaussian distribution;
\item \textbf{SHDC}: Does not conform to any common distributions and manifests smaller peaks at both extremes, corresponding to overexposed or underexposed pixels; 
\item \textbf{SHAC}: Conforms closely to a Laplace distribution.
\end{itemize}

We also conduct an empirical study to verify the agreement between the Gaussian attributes and the associated probability distributions mentioned above. 
Particularly, Shannon entropy~\cite{entropy} of quantized Gaussian attributes is calculated based on their histogram and compared to the actual bits per sample (bps) achieved by our estimation, as shown in Tab.~\ref{tab: EntropyGS_entropy}. Since Shannon entropy represents the theoretical minimum code length for lossless coding, from Tab.~\ref{tab: EntropyGS_entropy}, we see that by parameterizing the distributions of Gaussian attributes with estimated distributions, trivial coding overhead can be achieved.
This observation demonstrates the near-optimal precision of applying commonly used distributions to characterize the Gaussian attribute statistics.

\begin{table}[h!]
\centering
\resizebox{0.4\textwidth}{!}{
\begin{tabular}{ccccc}
\toprule
\textbf{Attribute} & \textbf{Q} & \textbf{Entropy} & \textbf{Actual} & \textbf{Overhead (\%)} \\
\midrule
Rotation  & 8 & 6.243 & 6.261 & 0.29 \\
Scaling   & 8 & 6.769 & 6.775 & 0.09 \\
Opacity   & 8 & 6.337 & 6.342 & 0.08 \\
SHAC     & 4 & 2.436 & 2.446 & 0.41 \\
\bottomrule
\end{tabular}
}
\caption{Verify the agreement between the true data distribution and estimated distribution. The results are averaged on Mip-Nerf 360, Q means quantization depth (discussed in Sec 4.3)}
\label{tab: EntropyGS_entropy}
\vspace{-0.3cm}
\end{table}

These statistical insights are foundational to our compression method. The strategy is straightforward: for rotation, scaling, opacity, and SHAC, we estimate their distribution parameters using maximum likelihood estimation (MLE) or expectation-maximization (E-M). 
For Geometry and SHDC, we combine them as a point cloud with color attributes, followed by applying an off-the-shelf point cloud compression method to it.

\subsection{Quantization Sensitivity}
\label{sec:sensitivity}
Our entropy coding method relies on quantization to obtain discrete representations. Thus, it is essential to investigate each attribute group's sensitivity to quantization errors. Please note that the \emph{quantization} used in this work is for entropy coding only, as defined in information theory. It is different from ``quantization'' mentioned in previous works where it indicates the reduced precision for neural network parameters and computation.

We examine the impact of quantization across various Gaussian attributes by analyzing the rendering degradation relative to error magnitude (see Fig.~\ref{fig: EntropyGS_sense}). Normalized quantization error is calculated by scaling the mean-squared error (MSE) by the average value of each attribute group.
Our study reveals substantial variability in quantization sensitivity: geometry is the most sensitive while SHAC exhibits the least sensitivity. This insight inspires an adaptive quantization strategy to optimize the rate-distortion tradeoff. The approach is straightforward: apply finer quantization steps to more sensitive attribute groups and coarser steps to less sensitive ones.

Furthermore, given that geometry is highly sensitive while SHAC is the least sensitive and yet consumes the most memory (76.3\%), 
they deserve additional treatment.
As to be seen in the next section, for geometry, we focus on reducing sensitivity and minimizing PSNR degradation. For SHAC, we aim to achieve a smaller quantization depth to maximize memory savings. 

\begin{figure*}[t]
\centering
\includegraphics[width=1\textwidth]{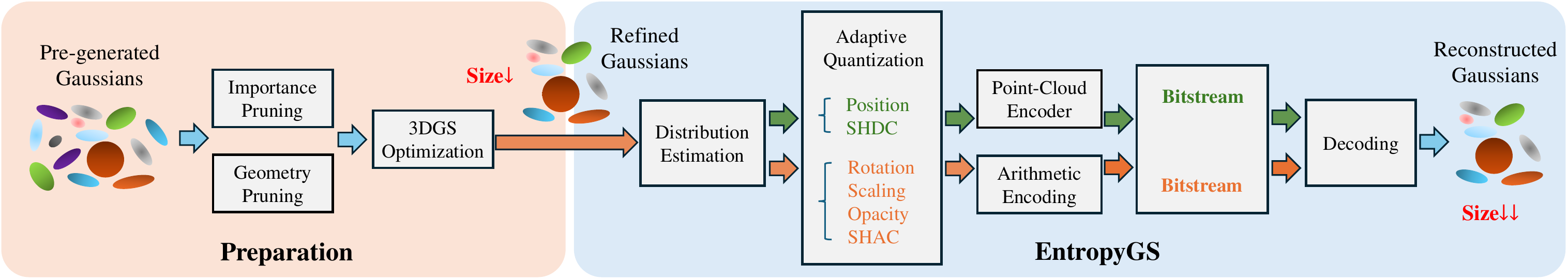}
\caption{Overview of our compression pipeline. Note that the complete decoding process is represented by one block in the figure due to space limit, the decoding process is symmetric with the encoding process. The first stage, Preparation, is optional in the pipeline. If no additional optimization is permitted in practice, the EntropyGS stage can be directly applied to the pre-generated Gaussians.}
\label{fig: Overview_prandft}
\vspace{-0.2cm}
\end{figure*}

\section{EntropyGS for Gaussian Splatting Coding}
\subsection{Overview of Coding Pipeline}
The overall coding pipeline consists of two distinct stages as illustrated in Fig.~\ref{fig: Overview_prandft}.

The pipeline is invoked with a \emph{preparation} stage, that is shared with earlier approaches. Here, the pre-generated Gaussian model undergoes refinement through statistic-guided pruning and subsequent optimization to get more compression-friendly Gaussians.

The second stage, our primary contribution called \emph{EntropyGS}, constitutes the core of the compression process. EntropyGS performs adaptive quantization and entropy coding, leveraging statistical distribution estimations.

It is important to highlight that EntropyGS is designed as an independent plug-in module. Its operation is not limited by the design of the preparation stage, allowing it to be flexibly applied to pre-generated 3DGS models with or without prior pruning and optimization.

\subsection{Preparation with Pruning and Optimization}
\label{sec: EntropyGS pruneandfinetune}
 During the preparation stage, redundant Gaussians are removed by pruning, and then the quantized Gaussians are updated to compensate for the rendering quality loss caused by pruning. We have adopted the same pruning strategy proposed by LightGaussian~\cite{LightGaussian} with small modifications to be presented next. 
In particular, the pruning is composed of two steps.
\begin{itemize}
\setlength{\itemindent}{1em}
    \item Pruning based on the importance level; and
    \item Pruning based on the geometry attribute.
\end{itemize}

The importance-based pruning step directly applies the approach from LightGaussian~\cite{LightGaussian} for its simplicity and effectiveness. Once the importance level of each Gaussian primitive is determined, we apply a threshold $\theta_1$(in \%) and remove less important Gaussians according to this value.

Driven by the observation from sensitivity analysis in~Section \ref{sec:sensitivity}, we propose a geometry-based pruning strategy to eliminate Gaussians that are hard to encode yet with little contribution to rendering. Specifically, Gaussians with geometry values far away from the scene center are pruned based on the following strategy: prune the top $\theta_2$ (in \%) of Gaussians with the highest absolute values for each direction $(x, y, z)$. Note that the $\theta_2$ is deliberately set small to avoid removing potentially meaningful Gaussians.

After pruning, the remaining number of Gaussians is:
\begin{equation} N_p = N_o \cdot (1 - \theta_1\%) \cdot (1 - \theta_2\%)^3 \end{equation}
where $N_o$ represents the number of initial Gaussians.

For post-pruning optimization, we first leverage SHAC Laplace distribution characteristics to rectify their values. 
This aims to scale down the range of SHAC values, thereby reducing the error introduced by subsequent min-max quantization. Please see the details in the supplementary.

Next, to compensate for the quality loss caused by pruning, an additional optimization step is required, adding approximately 20\% training overhead compared to the original 3DGS training. This overhead is considered acceptable for offline processing, especially when compared to the 170\% overhead introduced by joint methods like HAC. 

The preparation stage primarily serves as a test condition to ensure the Gaussians to be encoded in experiments are meaningful, which is a common practice in the literature. If additional optimization is not allowed, EntropyGS can still be applied to the pre-generated 3DGS.

\subsection{Factorized Entropy Coding with EntropyGS}
\label{sec: EntropyGS ec}
After the preparation phase, we compress the refined Gaussians with EntropyGS, which efficiently utilizes the statistical properties of Gaussian attributes.

We first quantized each group of Gaussian attributes with the quantization depth based on its sensitivity. The quantization is applied separately for each channel within one group (e.g. 4 channels in the Rotation group).

The quantization method applied to all attributes except for SHAC is a simple min-max uniform quantization:
\begin{equation} 
q(x) = \text{round} \left( \frac{(x - v_\textrm{min}) \cdot (L - 1)}{v_\textrm{max} - v_\textrm{min}} \right)
\end{equation}
where $x$ an entry of the channel to be quantized, $v_\textrm{min}$ and $v_\textrm{max}$ are the channel's minimum and maximum values, and $L$ is the number of quantization levels. Note that $L = 2\textsuperscript{Q}$ for a specific quantization depth $\textrm{Q}$.

We model the distributions of rotation, scaling, and opacity using Gaussian Mixture Models of up to four components. Note that this is also done separately for each channel within a group. The Expectation-Maximization (EM) algorithm~\cite{EM} is used to estimate the probability, mean, and variance for each component in an iterative manner. 
With the number of components $N$ limited to 4, the EM algorithm can be executed efficiently.

\begin{table}[h!]
\centering
\resizebox{0.3\textwidth}{!}{ 
\begin{tabular}{c c c c}
    \toprule
    & \multicolumn{3}{c}{\textbf{Quantization Depth}} \\
    \cmidrule(lr){2-4}
    \textbf{Attributes} & \textbf{Ours-L} & \textbf{Ours-M} & \textbf{Ours-S} \\
    \midrule
    Geometry & 17 & 16 & 15--16 \\
    Rotation & 8 & 8 & 7 \\
    Scaling & 8 & 8 & 7 \\
    Opacity & 8 & 8 & 7 \\
    SHDC & 8 & 8 & 8 \\
    SHAC & 4--5 & 3--4 & 2--4 \\
    \bottomrule
\end{tabular}
}
\caption{Quantization Depth for different coding configurations.}
\label{table:EntropyGSsetting2}
\vspace{-0.3cm}
\end{table}

For AC coefficients, Laplace distribution is used to model their values. We use maximum likelihood estimation (MLE) to determine the Laplace distribution parameters:
\begin{equation} 
\mu=\text{median($x$)},\quad b = \frac{1}{N} \sum\nolimits_{i=1}^N |x_i - \mu|
\end{equation}
where $\mu$ is the location and $b$ is the scaling parameter.
The Laplace nature of the AC coefficients is a significant and interesting finding.
Though out of the scope of this work, we hypothesize that this may stem from the L1-loss-based optimization used in 3DGS~\cite{3DGS}.

Following quantization and distribution estimation, we calculate the probability for each quantization interval to form the probability mass function (PMF). This PMF is then used by an Arithmetic Coding algorithm like~\cite{torchac} to achieve lossless compression.


\begin{figure*}[h!]
    \centering
    \begin{subfigure}[t]{0.35\textwidth}
        \centering
        \includegraphics[width=\linewidth]{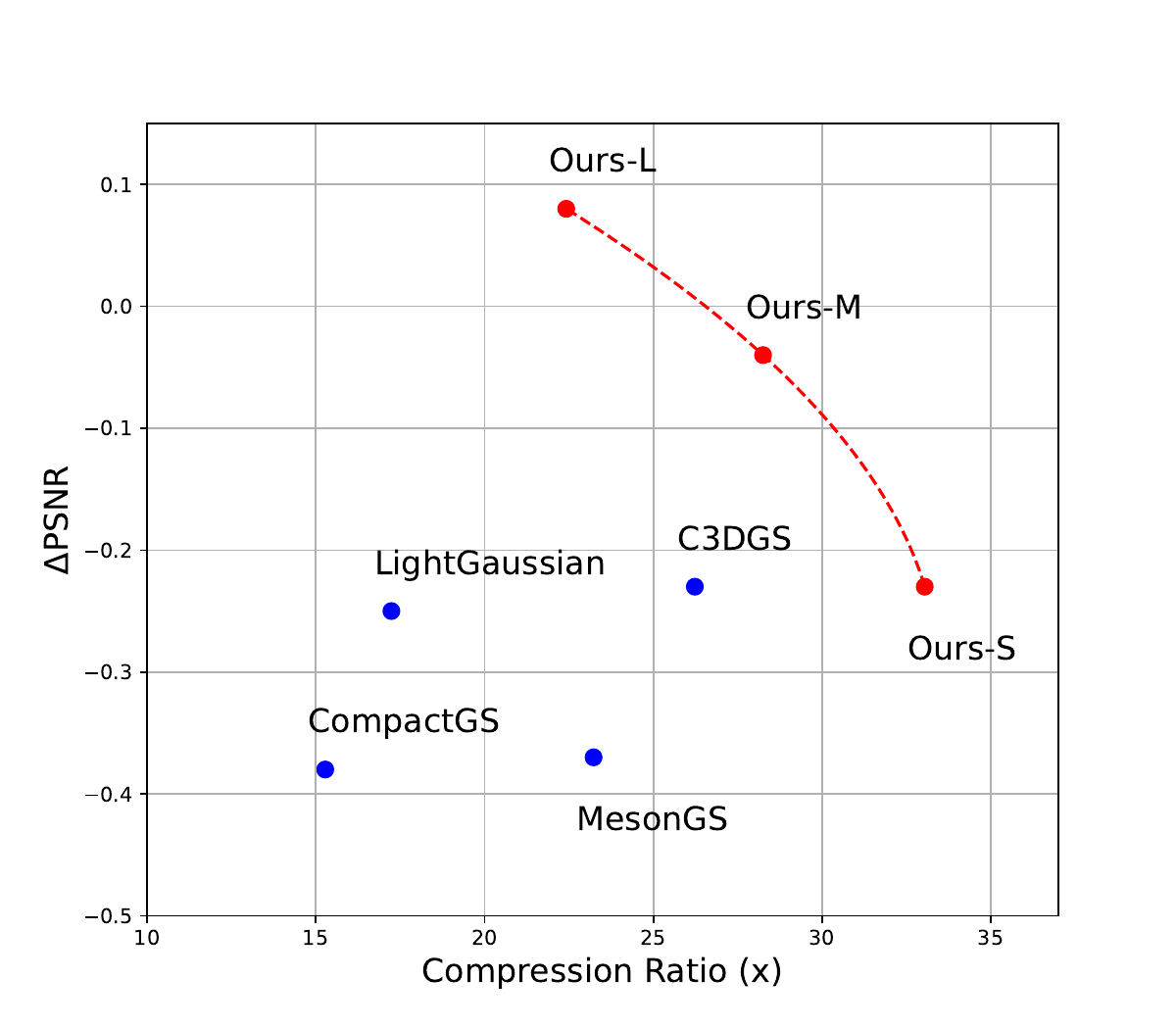}
        \caption{Comparison on Mip-NeRF 360 Dataset}
        \label{fig:mipnerf}
    \end{subfigure}
    \begin{subfigure}[t]{0.35\textwidth}
        \centering
        \includegraphics[width=\linewidth]{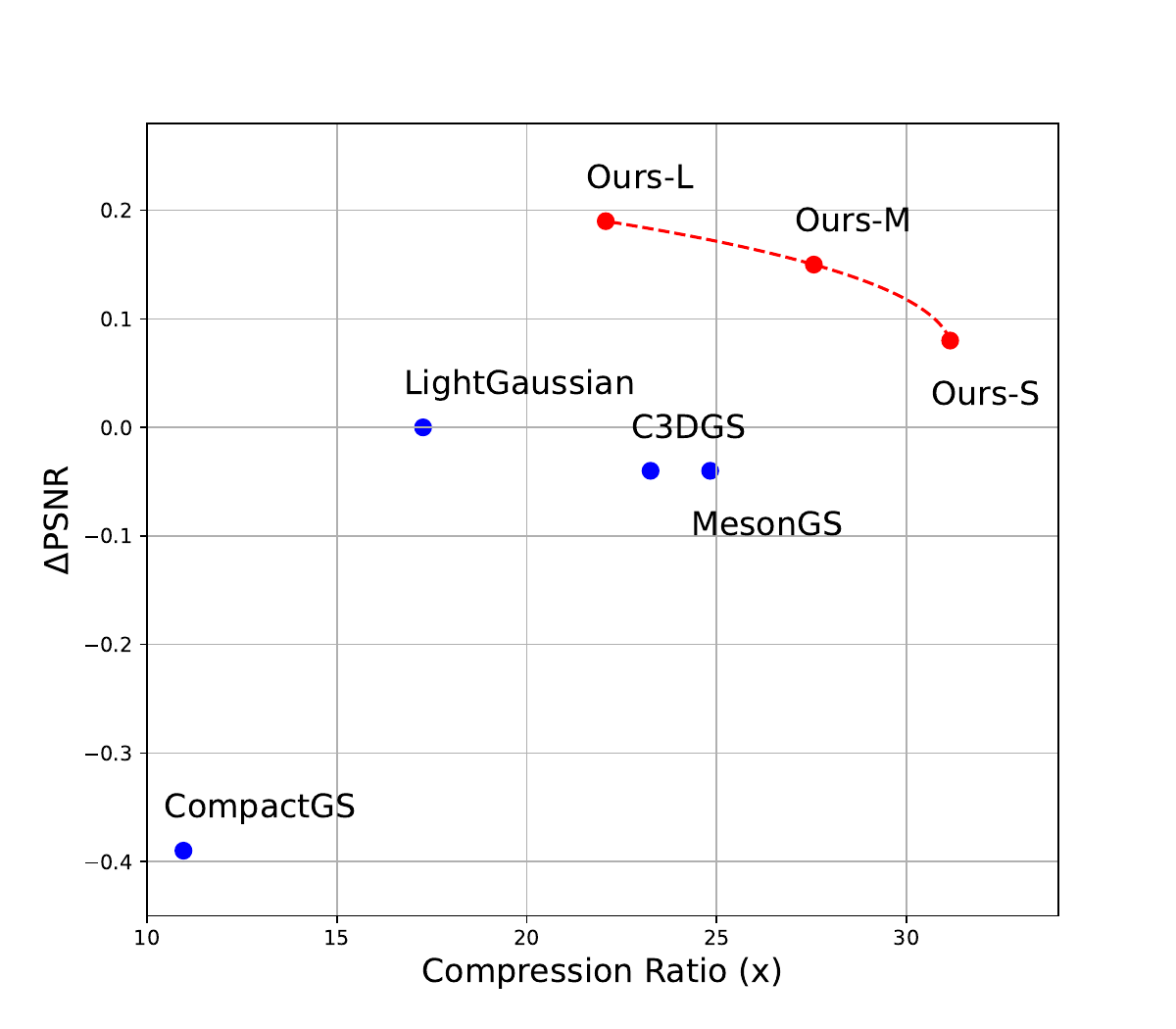}
        \caption{Comparison on Tank \& Temples Dataset}
        \label{fig:tank_temple}
    \end{subfigure}
    \caption{Performance comparison across datasets for different methods, \textbf{upper right is better}}
    \label{fig:EntropyGSmaincomp}
\end{figure*}

For the geometry and SHDC groups, we avoid fitting a commonly used distribution. Instead, we encode these attributes using a point cloud codec. Specifically, we employ the efficient G-PCC codec from the MPEG PCC standard\footnote{https://github.com/MPEGGroup/mpeg-pcc-tmc13}.
It treats the Gaussians with geometry and SHDC attributes as a color point cloud. Note a learning-based point cloud compressor can also be directly applied here. 

\subsection{Rate Control}
In our method, achieving different rates (model sizes) is straightforward and requires no additional training. It is achieved by simply adjusting the quantization depths applied to various groups of Gaussian attributes.

\section{Experimentation}
\subsection{Experiment Settings}
The proposed method is evaluated on Mip-NeRF 360~\cite{mipnerf}, Tank \& Temples~\cite{tanktemple}, and Deep Blending~\cite{deepblend}.

Our approach is compared against four recent 3DGS compression methods published in leading venues: LightGaussian~\cite{LightGaussian}, C3DGS~\cite{compressedgs}, CompactGS~\cite{compact3dgs}, MesonGS~\cite{mesonGS}. Recall the categorization of 3DGS coding illustrated in Section~\ref{sec:EntropyGS_relatedworks}, the selected methods are compression-specific methods that focus on the compression of pre-generated 3DGS scene without joint optimization, where heavy additional training is not feasible.
On the other hand, HAC~\cite{HAC} and SOG~\cite{SOG1}, as joint methods, have demonstrated excellent rate-distortion performance. 
However, these joint methods are not aligned with our scope, there they are not included for direct comparison (see supplementary for more information).

It is important to note that different baseline methods listed above have adopted different pre-generated 3DGS models, which results in variations in pre-compression metrics. For example, LightGaussian reported using a 3DGS with a PSNR of 27.53\,dB on the Mip-NeRF dataset, compared to the original 3DGS~\cite{3DGS} PSNR of 27.28\,dB. Therefore, to ensure a fair comparison, we focus on the metric \textbf{differences} between the compressed and uncompressed model results reported by each baseline method.
For instance, the PSNR difference is defined as:
\begin{equation} 
\Delta \text{PSNR} = \text{PSNR}_c - \text{PSNR}_o
\end{equation}
where PSNR$_c$ is the PSNR after compression and PSNR$_o$ is the PSNR before the compression.

The difference in PSNR (or SSIM/LPIPS) indicates the rendering quality loss due to compression. The interpretation of them follows the original metric's trend: for $\Delta$PSNR and $\Delta$SSIM, higher values mean better quality, while for $\Delta$LPIPS, lower values mean better quality. Typically, this difference (for PSNR/SSIM) is negative, representing a quality downgrade after compression. However, since additional optimization is applied, the difference can be positive, meaning the quality is improved.

\subsection{Implementation Details}

All our experiments were conducted on NVIDIA RTX 3090 and Intel 11700K, using the pre-generated models from the official repository of the 3DGS paper~\cite{3DGS} instead of retraining on our platform. For the hyperparameter setting, we categorize scenes into two groups: indoor and outdoor. The same processing strategy is applied to all scenes within each group, demonstrating the adaptivity of our method, i.e., no scene-specific configurations are required. See the settings for the pruning and optimization stage in the supplementary.

Based on the described rate control method, we introduce three coding configurations of EntropyGS: EntropyGS-L, EntropyGS-M, and EntropyGS-S, corresponding to large, medium, and small bitstream sizes, respectively.
See Tab.~\ref{table:EntropyGSsetting2} for details.
All configurations are applied to the same 3DGS output from the preparation stage.

\begin{table*}[h!]
\centering
\resizebox{0.8\textwidth}{!}{
\begin{tabular}{ccccccccccccc}
\toprule
& \multicolumn{4}{c}{\textbf{Mip-NeRF 360}} & \multicolumn{4}{c}{\textbf{Tank \& Temples}} \\
\cmidrule(lr){2-5} \cmidrule(lr){6-9}
\textbf{Method} & $\Delta$\textbf{PSNR} $\uparrow$ & $\Delta$\textbf{SSIM}$\uparrow$ & $\Delta$\textbf{LPIPS}$\downarrow$ & \textbf{Ratio\textsuperscript{*}}$\uparrow$ & $\Delta$\textbf{PSNR} $\uparrow$ & $\Delta$\textbf{SSIM}$\uparrow$ & $\Delta$\textbf{LPIPS}$\downarrow$ & \textbf{Ratio\textsuperscript{*}}$\uparrow$\\
\midrule
3DGS~\cite{3DGS} & - & - & - & - & - & - & - & - \\
LightGaussian~\cite{LightGaussian} & -0.25 & \textbf{-0.005} & 0.016 & 17.24$\times$ & \underline{0} & -0.005 & 0.012 & 17.27$\times$ \\
C3DGS~\cite{compressedgs} & \underline{-0.23} & -0.014 & 0.024 & \underline{26.23$\times$} & -0.04 & -0.009 & \underline{0.011} & 23.26$\times$ \\
MesonGS~\cite{mesonGS} & -0.37 & -0.009 & \underline{0.013} & 23.23$\times$ & -0.04 & \textbf{-0.001} & \textbf{0.006} & \underline{24.83$\times$} \\
CompactGS~\cite{compact3dgs} & -0.38 & -0.014 & 0.025 & 15.28$\times$ & -0.39 & -0.014 & 0.023 & 10.96$\times$ \\
\midrule
EntropyGS (Ours)-M & \textbf{-0.04} & \underline{-0.007} & \textbf{0.012} & \textbf{28.25$\times$} & \textbf{0.15} & \underline{-0.005} & 0.013 & \textbf{27.56$\times$} \\
\bottomrule
\end{tabular}
}
\caption{Comparison of methods on Mip-NeRF 360 and Tank \& Temples Datasets. The best results overall are \textbf{bolded} in each metric, and the second-best results are \underline{underlined}. Our method achieves a significantly higher PSNR while offering the highest compression ratio. \textsuperscript{*}The ratio denotes compression ratio (the original model size divided by the model size after compression)}.
\label{table:EntropyGS_table2}
\vspace{-0.3cm}
\end{table*}

\begin{table}[h!]
\centering
\resizebox{0.45\textwidth}{!}{%
\begin{tabular}{ccccc}
\toprule
& \multicolumn{2}{c}{\textbf{CPU coding time}} & \multicolumn{2}{c}{\textbf{GPU coding time}} \\
\cmidrule(lr){2-3}\cmidrule(lr){4-5}
\textbf{Method} & \textbf{Encoding} & \textbf{Decoding} & \textbf{Encoding} & \textbf{Decoding} \\
\midrule
EntropyGS (Ours)-M & 16.4s & 13.8s & 3.6s & 0.2s \\
\bottomrule
\end{tabular}
}
\caption{CPU and GPU coding time averaged on Mip-NeRF 360}
\label{table:EntropyGS_table3}
\vspace{-0.3cm}
\end{table}

\subsection{Main Results}

In Tab.~\ref{table:EntropyGS_table2} and Fig.~\ref{fig:EntropyGSmaincomp}, we evaluate our method alongside other 3DGS compression methods, note that in the table, we only show the result of EntropyGS-M to facilitate the comparison. Experiments were conducted on the Mip-NeRF and Tank \& Temples datasets, as some baseline methods do not report results for the Deep Blending dataset.

Our method achieves a significantly smaller model size across all datasets, effectively reducing storage requirements. Despite this compression, EntropyGS maintains comparable or improved PSNR values, particularly in the Large configuration, which shows minimal visual quality degradation. The flexibility of the rate control method enables a tradeoff between compression and rendering quality, making our approach ideal for applications that demand flexibility.

Comparisons with these baselines indicate that EntropyGS consistently outperforms them on these two benchmarks. For example, it achieves both higher $\Delta$PSNR and compression ratio compared to all baseline methods. The competitive performance of EntropyGS demonstrates the effectiveness of our proposal, which combines distribution estimation, adaptive quantization, and entropy coding, 
guided by an analysis of 3DGS properties.

Furthermore, EntropyGS achieves competitive performance with a streamlined, low-complexity design, unlike baseline methods that require trainable modules such as the codebook for vector quantization or neural networks for probability estimation. This streamlined design provides flexible and transparent control over the trade-off between rendering quality and compression ratio and can be easily optimized for specific hardware configurations. 

In terms of encoding and decoding complexity, it is worth noting that only two methods, C3DGS (5 minutes) and MesonGS (1 minute), have reported encoding times in their respective papers on GPU platform. We report both the encoding and decoding time in Tab.~\ref{table:EntropyGS_table3}. As expected, our method demonstrates a significantly faster encoding and decoding time, which is due to the factorized entropy model design and learning-free statistical distribution estimation.

\subsection{Ablation Studies}

\begin{table}[h]
\centering
\resizebox{0.4\textwidth}{!}{
    \begin{tabular}{ccc}
        \toprule
        \textbf{Method} & \textbf{PSNR} & \textbf{Size (MB)} \\
        \midrule
        Original 3DGS & 27.28 & 760.1 \\
        + Importance Pruning & 25.54 & 266.0 \\
        + Geometry Pruning & 25.45 & 264.4 \\
        + Optimization & 27.42 & 264.4 \\
        + Quantization & 27.24 & 41.5 \\
        + Entropy Coding & 27.24 & 32.2 \\
        + G-PCC & 27.24 & 26.9 \\
        \bottomrule
    \end{tabular}
}
\caption{Ablation Study: how each step in the pipeline influences the rendering quality and size of 3DGS, tested on Mip-NeRF360}
\label{table:EntropyGSab1}
\vspace{-0.3cm}
\end{table}

In this section, we examine the effectiveness of each step proposed in the method and how they contribute to the final performance (See Tab.~\ref{table:EntropyGSab1}), in this experiment, the EntropyGS-M setting is used.

The pruning of 3DGS primitives initially reduces model size while causing some rendering quality loss. However, the subsequent post-optimization effectively restores the quality and even slightly improves PSNR. Our proposed geometry pruning has minimal immediate impact, as only a small percentage of Gaussians are pruned, but this step helps lower quantization errors in later stages. 

Applying adaptive quantization to different attribute groups within 3DGS further reduces the model size, though it introduces some rendering quality degradation due to quantization error. The following entropy coding compresses the quantized values further in a lossless manner. Entropy coding is not applied to the geometry or SHDC attributes. Instead, we use the standard G-PCC point cloud codec in lossless mode to encode these components, achieving the smallest model size. Notably, the combination of quantization, entropy coding and point cloud coding alone achieves a 10× compression ratio without requiring any additional training (e.g., preparation stage mentioned before). This highlights the practical benefits of our proposed EntropyGS, which reduces the 3DGS model size by 10× while enabling fast encoding and decoding with no further optimization.

\section{Conclusion}
Motivated by a novel insight into the statistical characteristics of 3D Gaussian Splatting attributes, we introduce a simple yet effective compression scheme tailored for coding 3D Gaussian Splatting models. Experimental results demonstrate that our method achieves competitive compression performance while remaining both straightforward and computationally efficient.
It is important to note that EntropyGS adopts a factorized entropy model, encoding each attribute channel independently. This design offers advantages in simplified probability estimation, reduced computational overhead, and enhanced scalability. While rate-distortion performance could be further improved by accounting for the underlying relationships between attributes other than SHAC, efficiently incorporating them is a non-trivial challenge, which we leave for future work.

{
    \small
    \bibliographystyle{ieeenat_fullname}
    \bibliography{main}
}

\end{document}